\documentclass[preprint,5p,twocolumn]{elsarticle}
 \usepackage{multirow}

\usepackage{xspace}
\usepackage{lineno,hyperref}
\modulolinenumbers[5]

\usepackage{todonotes}
\usepackage[utf8]{inputenc}
\journal{Pattern Recognition Letters}

\usepackage{amsmath}
\usepackage{booktabs}
\usepackage{float}
\usepackage{placeins}

\usepackage{xcolor}

\usepackage[compact]{titlesec}         
\titlespacing{\section}{2pt}{2pt}{2pt} 

\begin{document}

\begin{frontmatter}

\title{MSdocTr-Lite: A Lite Transformer for Full Page Multi-script Handwriting Recognition}


 \author[firstaddress,secondyaddress,thirdaddress]{Marwa Dhiaf\corref{mycorrespondingauthor}}
 \cortext[mycorrespondingauthor]{Corresponding author}
 \ead{marwa.dhiaf.doc@enetcom.usf.tn}

 \author[firstaddress]{Ahmed Cheikh Rouhou}
 \ead {a.cheikhrouhou@instadeep.com}
 \author[secondyaddress,thirdaddress]{Yousri Kessentini}
 \ead {yousri.kessentini@crns.rnrt.tn}
\author[firstaddress]{Sinda Ben Salem}
 \ead {s.bensalem@instadeep.com}
\address[firstaddress]{InstaDeep}
 \address[secondyaddress]{Digital Research Center of Sfax, Sfax 3021, Tunisia}
 \address[thirdaddress]{SM@RTS : Laboratory of Signals, systeMs, aRtificial Intelligence and neTworkS}

 \begin{abstract}
 The Transformer has quickly become the dominant architecture for various pattern recognition tasks due to its capacity for long-range representation. However, transformers are data-hungry models and need large datasets for training. In Handwritten Text Recognition (HTR), collecting a massive amount of labeled data is a complicated and expensive task. In this paper, we propose a lite transformer architecture for full-page multi-script handwriting recognition. The proposed model comes with three advantages: First, to solve the common problem of data scarcity, we propose a lite transformer model that can be trained on a reasonable amount of data, which is the case of most HTR public datasets, without the need for external data. Second, it can learn the reading order at page-level thanks to a curriculum learning strategy, allowing it to avoid line segmentation errors, exploit a larger context and reduce the need for costly segmentation annotations. Third, it can be easily adapted to other scripts by applying a simple transfer-learning process using only page-level labeled images. Extensive experiments on different datasets with different scripts (French, English, Spanish, and Arabic) show the effectiveness of the proposed model.
 \end{abstract}
 

 \end{frontmatter}

\section{Introduction}

Handwritten Text Recognition (HTR) aims to transform scanned handwritten documents into machine-encoded text. HTR is still a hard problem due to the high variance of writing styles, illegible handwriting, poor quality, and degradation. Furthermore, each handwritten script has specific properties (writing direction, character shapes, ligatures, etc) making the problem more challenging. 

 Current HTR models are leading to an acceptable performance \cite{ref_article6}, especially for modern documents with legible handwriting styles, known languages, vocabulary, and syntax. However, most of these systems rely on a segmentation phase before the recognition task. The segmentation consists of partitioning the document images into homogeneous components (characters, words, or lines). Early HTR approaches have applied segmentation techniques to extract character-level images from documents. These techniques suffer from a deficiency in performance accuracy as the segmentation generally fails due to the cursive and unconstrained nature of the handwriting. To {overcome} this problem, word-based segmentation methods have been introduced, where word images are extracted from the document before feeding {them} into an HTR system.  But, this kind of approach faced some problems due to the irregularity of the inter-space and the intra-space between words. 

In the last decade, word segmentation-based approaches \cite{kes09} were abandoned in favor of text line recognition \cite{ref_article2,kes18, bluche2016joint}: a document can be segmented into individual lines to surpass any inconsistency between words and have more context. This kind of approach generally achieves state-of-the-art performance for mostly all types of documents (modern, historical, multi-script, etc) \cite{marti2002iam,grosicki2011icdar}, without suffering the heavy burden of character/word segmentation. Despite their success, line segmentation is also challenging due to non-uniform text line skew/slant or closely situated and touching text lines. Such problems can considerably affect the performance at the recognition stage. 
With the recent success of deep learning, the HTR community started exploring handwriting recognition at the paragraph or page level to avoid any intermediate segmentation pre-processing step. In \cite{bluche2017scan} an architecture for paragraph-level HTR with the use of { Multi-Dimensional Long Short-Term Memory Recurrent Neural Networks (MDLSTM) } and the attention mechanism was proposed. However, they later abandoned this approach due to high memory requirements, the lack of GPU acceleration for the training of MDLSTM, and intractable inference time.

Other approaches \cite{singh2021full,coquenet2022dan} have {developed sequence-to-sequence} (Seq2Seq) models, based on the transformer \cite{vaswani2017attention} architecture, for page-level handwritten-document text recognition. These systems have reported state-of-the-art performances compared to line-level HTR ones. Nevertheless, training such models requires a huge amount of annotated data. Moreover, the length of extracted feature sequences, compared to line-level approaches, can be very large resulting in weeks of training consuming an  {immersive} amount of GPU resources. Finally, porting these models to low-resource languages can be a difficult task as it requires a huge quantity of data to retrain the model from scratch.

Motivated by the above observations, we propose a lite transformer model for page-level handwritten document image recognition. The proposed model involves a limited number of parameters and can be trained without the use of external data. Our system is trained with a curriculum learning strategy, allowing it to learn the reading order and to scale to large input text images. This curriculum learning strategy is performed only once and the resulting model will serve as a starting point to train models with real page-level data even on different scripts. The proposed architecture can be trained using standard GPUs as it requires lower memory allocation, compared to similar proposed systems.
The major contributions of this paper can  be therefore summarized as follows:

\begin{itemize}
    \item We propose an end-to-end lite transformer-based model for handwritten text recognition. The proposed model works at {the} page level to avoid unrecoverable early line segmentation errors, exploit a
    larger context and reduce the need for costly segmentation annotations.
    
    \item {We propose to use a curriculum learning strategy } that allows the model to be trained using a limited amount of annotated data and can learn the reading order at the page level. 
    
     \item We demonstrate that our lite transformer, built with only two layers and a single head of self-attention, can be easily adapted to other scripts by applying a simple transfer-learning process using only page-level labeled images. 
    
    \item Extensive comparative experiments are conducted to validate the effectiveness of our approach. We have tested our model on multiple scripts and languages, including English, French, Spanish, and Arabic. The obtained results confirm the effectiveness of the proposed model.
\end{itemize}

The remaining parts of this paper are structured as follows. Related works are presented in Section 2. Section 3 describes the proposed approach. Section 4 displays the experimental results. The conclusion and perspectives are stated in section 5.

\section{Related works}\label{sec2} 
\subsection{Line-level HTR systems}\label{subsec1}

In the literature, using line-level segmented images to perform handwritten text recognition has gained a lot of interest. Systems following this approach benefit from the minor segmentation effort compared to the word or character-level systems. 

With the rise of deep learning, many deep neural networks have been proposed based mostly on the recurrent architectures \cite{graves2008offline,bluche2016joint}. However, such models require a large amount of annotated data to be trained. 
To this end, authors in \cite{ref_article2}, proposed data augmentation and normalization techniques using Convolution Neural Networks combined with Long-Short-Term Memory CNN-LSTM, which significantly reduces the error rate on handwriting recognition tasks. Having unlimited annotated data, in turn, does not alleviate the issue of parallelization during the training stage. They suffer from the lack of computation parallelization inherently due to recurrent layers, which impact both training and inference time. Consciously, the next methods have focused on training systems without a recurrent process.

One can notice some latest trends, based on transformer architecture, in which, authors in \cite{ref_article6} propose an end-to-end model, which dispenses any recurrent network for HTR of text-line images. The architecture is based on multi-head self-attention layers
both at the visual and textual stages. Such a model aims to tackle both the proper step of character recognition from images, as well as to learn language-related dependencies of the character sequences to be decoded. The proposed work reaches competitive state-of-the-art performance at the line level.
In brief, line-level methods require segmentation and throwing away useful context.

\subsection{Page level HTR systems}\label{subsec2}
To avoid unrecoverable errors of segmentation, and to exploit larger context information,  some approaches are moving from line-level recognition to paragraph-level and even page-level recognition.

Methods like \cite{bluche2016joint}, \cite{bluche2017scan} applied an encoder-decoder with attention mechanisms to the task of multi-line text recognition. Both architectures use a CNN+MDLSTM encoder and an MDLSTM-based attention module. The encoder produces feature maps from the input image while the attention module recurrently generates line or character representations applying a weighted sum between the attention weights and the features. Finally, the decoder outputs character
probabilities from this representation. Both those two approaches present interesting implicit segmentation reaching competitive results. Nonetheless,  they are computationally expensive due to the MDLSTM layers, which lead to {long} training and inference time. 
{In addition to the complexity of the MDLSTM convolutions, the MDLSTM-based attention needed to be recomputed for each predicted character. }

Other approaches \cite{singh2021full,rouhou2022transformer} proposed different
models which are based on an attention mechanism to predict the text in a character-by-character way. Those are trained with the cross-entropy loss and using a special end-of-transcription token. Authors of  \cite{ rouhou2022transformer} used a  transformer to evaluate their approach to paragraph-level images. \cite{singh2021full} introduced an encoder-decoder architecture, using a {Residual Networks (ResNet) }\cite{7780459} for encoding the image, and a transformer \cite{vaswani2017attention} for decoding the encoded representation into text. This model is trained to recognize, in addition to the text, the presence of non-textual areas (such as tables or drawings). This model has been trained to recognize full pages of handwritten or printed text without image segmentation. Despite the good results achieved using this configuration, this model is data-hungry and requires a lot of memory and computation resources.

In \cite{coquenet2022dan}, a new end-to-end segmentation-free architecture was created to address document image recognition in terms of both text and layouts. They proposed Document Attention Network  (DAN), an end-to-end transformer-based model for handwritten document recognition, including the textual components and the logical
layout information. { The model has been evaluated using two different datasets where competitive results have been reported at page and double-page levels. However, the main drawback of this work is that the model architecture is very large and requires a huge quantity of data for the training phase. In addition,  the model involves  an auto-regressive attention-based   character-level prediction process, which leads to high prediction times (a few seconds per document image).}

Considering the {limitations} of previous works, we propose in this paper MSdocTr-Lite, an end-to-end lite transformer-based model for multi-script handwritten document recognition.

\section{ Proposed approach } 
\label{section:sec3}

\begin{figure*}
\centering
\includegraphics[scale=0.70]{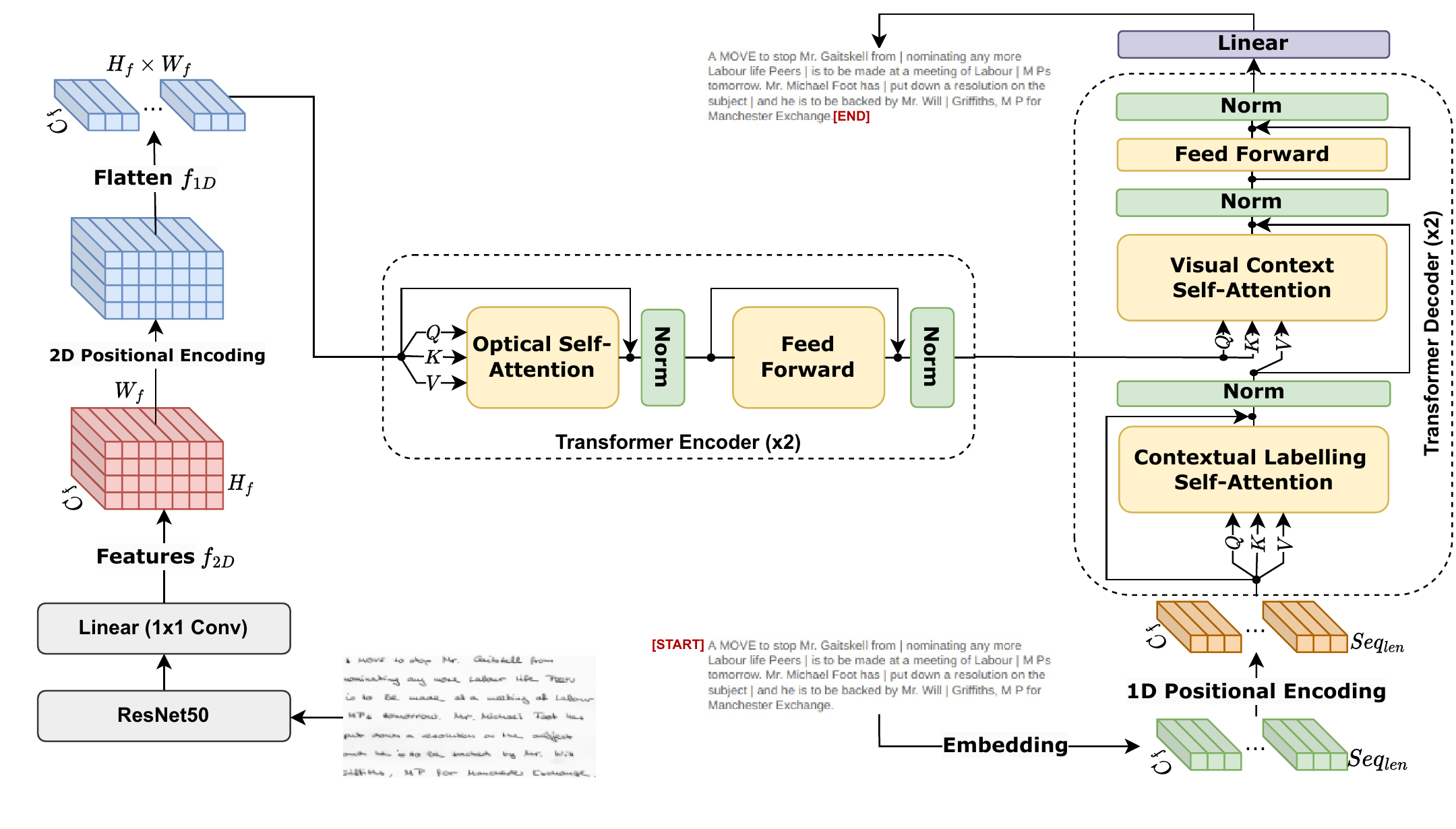}
\caption{Overview of the proposed architecture. Our lite transformer is composed of a transformer encoder combining convolutional layers and transformer layers and of a transformer decoder.} \label{fig1}
\end{figure*}
The proposed model is depicted in Figure \ref{fig1}. It is composed of two parts: a Transformer-Encoder and a Transformer-Decoder, intending to map an inputted image that contains text into a sequence of characters, i.e, an {image-to-sequence} architecture. \cite{vaswani2017attention}.

\subsection{Transformer-Encoder  }\label{subsubsec4}
The main objective of the encoder is to extract high-level feature representations from the sequence of features and encode the visual information. This component is mainly composed of a backbone and 2 transformer layers.  The backbone consists of the feature extraction architecture to represent the input image into high-dimensional vectors. It uses the ResNet-50 \cite{7780459} architecture without its last two layers( the average pool and linear projection). The ResNet produces a  feature representation of the full-page image. {A 2D-convolutional layer (Conv2D) with a kernel size of (1 $\times$ 1) is added, to match the number of features from the backbone network and the transformer input. }After, the 2D positions encoding \cite{9151002} are merged with these features to add positional information to each feature vector. Next, the 2D features map is transformed into 1D sequential features using {tensor reshaping flattening operation}. At this level, the output of the flattened layer matches the requirements of the transformer layers. After performing these steps on a page-level image, we obtain a {feature} sequence of the image. The feature sequence is then passed to a stack of 2 {transformer} layers (as shown in Figure \ref{fig1}).

\subsection{Transformer-Decoder}\label{subsubsec5}
The decoder is similar in structure to the transformer layers of the encoder except that it includes a masked language model. The masked language modeling approach is introduced to model the relation of the characters that are output from the decoder. It learns the correlation between visual encoded features and their corresponding characters. There are two types of symbols in our approach: visual labels and contextual labels. The visual labels ($V_{labels}$ ) are the readable characters from texts in images. The contextual labels ($C_{labels}$)are the start of page ($<sop>$), the end of line ($<eol>$), and the end of page ($<eop>$). The number of classes ($NB_{class}$) of the output layer includes the number of visual labels ($V_{labels}$ ) and the contextual labels ($C_{labels}$). {We used the teacher forcing strategy during the training, which means that the model uses the ground truth as input instead of model outputs from prior time steps}. The input of the decoder is the textual contents of the image surrounded by $<sop>$ and $<eop>$ tags. After preparing the input text, the characters are embedded into a  vector of $h$ values as mentioned in equation \ref{eq8}.
\begin{equation}
\label{eq8}
Embed=Embedding(NB_{class},h)
\end{equation}
Then, the final input of the transformer decoder ($F_{labels}$) is prepared by applying 1D Positional encoding \cite{10.5555/3295222.3295349} to the embedded labels.  
\begin{equation}
\label{eq10}
F_{labels} = 1DPE(Embed(labels))
\end{equation}
The shape of $F_{labels}$ is $(h, N)$ where $N$ represents the length of the input labels sequence.  
\subsection{The curriculum learning strategy} 
\label{section:my}


  {
 Transformers are data-hungry models and need large datasets for training. In the ideal case where a huge quantity of full-page document images is available, it was proven in \cite{singh2021full} that a large transformer model can be trained without the need for a curriculum learning strategy. In our case, where only a limited amount of training data is available, the lite transformer model fails to converge (as described in table 2) without an adapted learning strategy. 
For this aim, we propose a curriculum learning strategy that involves 3 stages: in the  first stage, the model is trained using a generated dataset composed of 50k small block images, where each block is composed of a minimum of 1 line and a maximum of 4 lines and each line contains from one to four words. In the second stage, the model is fine-tuned on a more complex dataset containing 78k blocks where the number of lines  arbitrarily varies between 3 and 14. The aim of this second pre-training stage is to improve the capacities of the model to deal with longer sequences corresponding to more complex documents (with a higher number of lines). These two pre-training steps are performed only once, they can be done using synthetic data, or using any public dataset providing the segmentation annotation (which is done in our case using the IAM dataset). The resulting model will serve then as a starting point for the third training stage where the fine-tuning is performed using the original page-level document images of the target dataset (even when the training is performed on a different script as explained in section 3.4). 
We note that during the inference, the model does not need any line or other lower-level segmentation step, and the recognition is performed at the page level. We {also note} that several data augmentation techniques are applied in all these previous training stages including random rotation,  brightness, contrast, perspective, and Gaussian noise. Zero padding is used to create mini-batches.
}

\subsection{Transfer Learning }\label{sec4}

In this section, we investigate the use of transfer learning to improve the performance of the transformer model {in the context of low-resource multi-script handwriting recognition}. 

{We propose a simple, yet effective, transfer learning approach. The idea here consists in using the trained model described in section 3.3 as a starting point and improving its generalization ability on other scripts. Since the model has already learned the reading order on page-level documents {thanks} to the curriculum learning strategy, the fine-tuning {of} new scripts is done on a single stage using the  target script page-level document images. This presents several benefits like saving training time and not needing a lot of data. } 

{ Fine-tuning a transformer-based model consists in adapting it to learn the knowledge of both the visual feature extraction (optical model) and the language model. Fine-tuning the optical model consists in training the backbone and the encoder layers to reduce the domain shift between the source and target domains. In fact, images from each domain contain different distributions resulting from irregularities in terms of line spacing, different handwriting skew/slant, and different writer styles... 
The second aspect of transfer learning concerns the fine-tuning of the decoder layers to adapt the language model to capture the inherent nuances of the target text script.  Next, the visual-context self-attention is updated to match a newly learned representation of the encoder and the language model. Finally, the last linear layer is replaced with a new one, with random weights, to match the number of characters of the target script.
We present in the experimental section an ablation study exploring the impact of fine-tuning on the different components of the transformer and its effect on the model performance. We find that the fine-tuning of both the optical model and the language model {gives} the best results.    
}

\section{Experiments}\label{sec6}

This section is dedicated to {presenting} the evaluation of the lite transformer in the context of multi-script handwriting recognition. We start by presenting the four used datasets: IAM \cite{marti2002iam}, RIMES \cite{grosicki2011icdar}, KHATT \cite{mahmoud2012khatt}, and Esposalles \cite{ROMERO20131658}. We present {an} ablation study to evaluate the proposed curriculum learning strategy and the transfer learning process. Finally, we compare our results to the state-of-the-art approaches.

\subsection{Datasets description}\label{subsec11}

\begin{itemize}
  \item   {
  IAM dataset contains forms of handwritten English text created by 657 writers. The training set comprises 747 documents (6,482 lines, 55,081 words), the validation set is composed of 116 documents (976 lines, 8,895 words) and the test set contains 336 documents (2,915 lines, 25,920 words).}

\item  {RIMES is a popular handwriting dataset composed of gray-scale images
of French handwritten text taken from scanned {mail}. The images have a resolution of 300 dpi. In the official split, there are 1,500 pages for training and 100 pages for evaluation. Segmentation and transcription are provided at the page, line, and word levels. We kept the same split between training, validation, and testing at a page level.}

\item { KHATT is a freely available Arabic handwriting dataset. It consists of scanned handwritten pages with different writers, text, and resolutions. Pages segmentation into lines is also provided to allow the direct evaluation of recognition systems without layout analysis. We have used 1400 blocks for training, 300 blocks for validation{,} and 300 blocks for testing.}

\item  { Esposalles is a public dataset proposed in ICDAR 2017 competition on Information Extraction from Historical Handwritten Records (IEHHR) ~\cite{8270158}. The dataset contains 125 Spanish handwritten pages, containing 1221 marriage records. We have used 872 records for training, 96 records for validation, and 253 records for testing.}
 \end{itemize}

\subsection{Implementations details}
 {The model was implemented using PyTorch and training was carried out using {the} Tesla T4 GPU. The lite transformer architecture is composed  of 2 transformer layers with one head attention  and a hidden size of 256. The large transformer is composed of 4 transformer layers with 8 heads and  a hidden size of 512. }
 { ADAM optimizer was utilized with a fixed learning rate of 0.0001. This learning rate is decreased to 0.00001 in the transfer learning stage.  We do not use any synthetic data, external language model, or lexicon constraints. To prevent over-fitting, regularization techniques are applied like a dropout of 0.1, data augmentation, and early stopping. }
 
{
We used the standard Character Error Rate (CER) to evaluate the quality of recognition. 
For a set of ground truths, CER is computed using the Levenshtein distance which is the sum of the character substitutions ($Sc$), insertions ($Ic$), deletions  ($Dc$), divided by the total number of the ground truth characters ($Nc$). The formula of this metric is given respectively as below:
\noindent
\begin{equation}
CER=\frac{Dc+Sc+Ic}{Nc}
\end{equation}}


\subsection{Evaluation of the curriculum learning strategy}\label{subsusec1}
In this section, we evaluate the interest of the proposed 3-stages curriculum learning strategy on the IAM dataset for both the lite and the large transformer architectures. {The obtained results are given in Table \ref{tab1}.}
 The curriculum learning strategy is carried out in three stages as described in Section \ref{section:my}. {The test set consists of the real pages of IAM with a resolution of ($1024\times1024$px)}.  In the first stage S1, the model was trained on small block images with a size of $256\times256$ pixels containing few lines and few words per line. The goal of this stage is to help the model learn the reading order on easy samples. A CER of $14.77\%$ was obtained due to the domain shift between the training and the test images in terms of resolution and quantity of text.  {In the second stage S2, the model was fine-tuned on higher resolution images ($1024\times1024$px)  containing an increased number of lines and words per line }. The aim of this second pre-training stage is to improve the capacity of the model to deal with longer sequences corresponding to more complex documents. A gain of $1.93$ in terms of CER is obtained compared to the model trained on stage 1. The third stage S3 consists of fine-tuning the model obtained in the previous stage on the real IAM dataset {using  the same image size as stage 2}. This last training stage improves the model performance and a gain of $6.44$
 in terms of CER is obtained compared to the model of the previous stage.

\begin{table}[!htp]
\centering
\caption{Evaluation of the training strategy in terms of CER depending on the transformer model size Lite/Large }\label{tab1}
\begin{tabular}{llllll}
\hline
S1 & S2 & S3 & FS & \begin{tabular}[c]{@{}l@{}} Lite \end{tabular} & \begin{tabular}[c]{@{}l@{}} Large \end{tabular} \\ \hline
x       &        &        &    & 14.77                                                          & 95.6                                                             \\
        & x      &        &    & 12.84                                                          & 43.16                                                            \\
        &        & x      &    & 6.4                                                            & 27                                                               \\
        &        &        & x  & 13                                                             & 40                                                               \\ \hline
\end{tabular}
\end{table}

To highlight {the} importance of the curriculum learning strategy, we have conducted another experiment where the model is trained from scratch (FS) on only one stage using all the training images used in the three stages. We obtained a CER of $13\%$ which confirms the interest of the curriculum learning strategy which helps the model to learn the reading order and clearly improves its convergence.
 Finally, we can observe that the performance of the lite transformer is better than the large transformer for all the training stages and scenarios. This confirms our choice to adopt a lite model in the context of low-resource handwriting recognition.


\subsection{Evaluation of the transfer learning process }
In this section, we report the evaluation of the transfer learning process considering the IAM dataset (English script) as a source domain and the RIMES dataset (French script) as a target domain. 
In this scenario, we initialized the model parameters by the weights of the original model trained on the IAM dataset, then we fine-tune just a subset of layers going from the last transformer layer to the first one.   We start by learning the FC layer and the embedding weights for the decoder while keeping the other parts of the model frozen. For example, $FC+embed, Decod{[}L1{]}$, in table \ref{table5}, indicates that the second layer of the decoder, the embedding weights, and the Fully connected layer are fine-tuned while keeping the rest of the layers frozen. Next, the first layer of the decoder (L0) is included in the set of layers to be re-trained. The remaining steps follow the same logic. An upper layer is added to the set of layers to be retrained until reaching the whole architecture. 

{The results of table  \ref{table5} correspond to a transfer learning from English to French handwritten text. We notice that by fine-tuning only the embedding layer and the FC layers, a CER of $30.7\%$ was obtained on the test set. By fine-tuning the decoder layers, an improvement of $5.3\%$ in terms of CER is obtained due to the adaptation of the language model, initially trained on English text, to capture the inherent nuances of the target text (French). We note here that English and French scripts share multiple similarities, notably most of their character sets and a number of true word cognates. This can explain why the obtained gain is not so high and it can be more important when the scripts of the source and the target domains present more dissimilarity. 
 The second aspect concerns the fine-tuning of the optical model, which includes the backbone and the encoder layers. As shown in Table \ref{table5}, a gain of $12.7\%$ in terms of CER is obtained which confirms that the transfer learning allows the model to reduce the domain shift between the source and target domains as IAM images  and RIMES images contain different distribution resulting from irregularities  in term of line spacing, different handwriting skew/slant and different writer styles. 
 }


\begin{table}[!htp]
 
\begin{center}
 
\caption{Evaluation of the transfer learning process on RIMES dataset}\label{table5}%

\begin{tabular}{ll}
\hline
Trainable layers                           & CER \\ \hline
FC+embed                                            & 30.7    \\
FC+embed,Decod{[}L1{]}                             & 37.9    \\
FC+embed,Decod{[}L1,L0{]}                            & 25.4    \\
FC+embed,Decod{[}L1,L0{]},Encod{[}L1{]}              & 17.3    \\
FC+embed,Decod{[}L1,L0{]},Encod{[}L1,L0{]}           & 15.6    \\
FC+embed,Decod{[}L1,L0{]},Encod{[}L1,L0{]},Backbone & 2.9     \\ \hline
\end{tabular}

\end{center}
\end{table}

From this ablation study, we can conclude that the best transfer learning strategy consists in fine-tuning all the model components. The model reaches a competitive result on the RIMES dataset as presented in section \ref{subsusecstate}. 

To explore the capacity of the model to be adapted to other scripts, additional experiments are conducted considering as a target domain the Esposalles dataset containing historical handwritten documents written in Spanish script. We note the size of images is  reduced for this dataset, to decrease the feature map sequence length and consequently speed up the convergence. A CER of $1.7\%$ is obtained on the Esposalles test set which confirms the capacity of the model to transfer the knowledge acquired from learning English handwritten text to recognize Spanish handwriting, even in the case of high domain shift between the source and the target datasets.     

The last and the most challenging experiment consists in fine-tuning the model trained on English handwriting to recognize Arabic handwriting (KHATT dataset). The challenge here consists in the fact that English and Arabic use completely different alphabets, and are naturally written in different directions (from right to left for Arabic and from left to right for English.    

To exploit the reading order knowledge acquired by the model on English script, we apply a horizontal flipping pre-processing to the images of the KHATT dataset before inferring them to the model. Additionally, the images are resized to $2048\times512$ as the maximum number of lines per image is 6. This helps the model to capture more contextual information and to map the high-level feature to the text of the ground truth during decoding. 
 {As reported in table \ref{khatt}, our model achieved competitive results on the KHATT dataset, which confirms that the model benefits from the knowledge acquired from the IAM dataset to learn the reading order at {a} record level, but is also able to improve its generalization ability to learn the specificity of the target script by transfer learning. }


\subsection{Comparison with state-of-the-art}\label{subsusecstate}

\begin{table*}[h]
\caption{Comparison with transformer models on the test set
of the IAM dataset. }\label{comp}
 
\centering 
\begin{tabular}{@{}lllllll@{}}
\toprule
\multirow{2}{*}{System}                    & \multicolumn{3}{l}{Architecture} & \multirow{2}{*}{\begin{tabular}[c]{@{}l@{}}Using \\ external data\end{tabular}} & \multirow{2}{*}{\begin{tabular}[c]{@{}l@{}}Number \\ of parameters  transformer\end{tabular}} & \multirow{2}{*}{CER} \\ \cmidrule(lr){2-4}
                                           & layers    & heads   & {hidden size}   &                                                                                 &                                                                                                 &                      \\ \cmidrule(r){1-1} \cmidrule(l){5-7} 
FPHR \cite{singh2021full}    & 6         & 4       & 260        & yes                                                                             & 6.3 M                                                                                           & 6.3                  \\
FPHR \cite{singh2021full}    & 6         & 4       & 260        & No                                                                              & 6.3 M                                                                                           & 6.7                  \\
DAN \cite{coquenet2022dan} & 4         & 8       & 256        & yes                                                                             & 7.6 M                                                                                           & -                    \\
Ours                                       & 2         & 1       & 256        & No                                                                              & 3.9 M                                                                                           & 6.4                  \\ \bottomrule
\end{tabular}
\end{table*}
In this subsection, we compare the performance of our model with the state-of-the-art models on different datasets. Note that our lite transformer model is evaluated without the use of an external language model and any lexicon constraint.

\begin{table}[H]
\centering
\caption{Comparison on the IAM dataset of the lite transformer with the state-of-the-art approaches.  }\label{compiam}
\begin{tabular}{@{}lll@{}}
\toprule
Type                             & Architecture                       & CER   \\ \midrule

\multirow{3}{*}{Paragraph level} & Bluche et al   \cite{bluche2017scan}  & 11.1  \\     
                                & Bluche et al   \cite{bluche2016joint}  & 7.9  \\     
                                 &{Bluche et al$^*$ }\cite{bluche2016joint}                        & {5.5}  \\ & {Yousef et al } {\cite{yousef2020origaminet}}                        &  {4.7 }  \\
                        
\\                                 
\multirow{3}{*}{Line level}     & Chowdhury  \cite{chowdhury2018efficient}                 & 8.10  \\
                                 & Bluche et al   {\cite{bluche2017scan}  }                    & 7.00  \\
                                 & Kang et al \cite{kang2020pay}               & 7.62 \\
  \\                               
Page level                        & Bluche et al \cite{bluche2017scan}                        & 16.2  \\
                                 & Carbonell \cite{carbonell2019end}  & 15.6 \\
                                 & FPHR \cite{singh2021full}                               & 6.7   \\
                                 & {Wigington et al$^*$ \cite{wigington2018start} }          &{ 6.4 }\\
                                 
                                 & Ours                                & 6.4   \\ \bottomrule
\end{tabular}\\
\footnotesize{$^*$ System using external language model}

\end{table}

Table \ref{compiam} provides the results on the IAM dataset. As one can notice, our lite transformer model surpasses line-based methods that use accurate localization information during training and {shows} competitive results compared to page and paragraph levels-based systems. 


In table \ref{comp}, we concentrate on the comparison with state-of-the-art approaches based on transformer  architecture. A special focus is given to the number of parameters in each model and to the kind of data used in the training stage. 
{ Models proposed in \cite{coquenet2022dan,singh2021full} have implemented larger transformer-based architectures, compared to our lite model. To train such models, a huge amount of data has been collected (up to 47 M text images in \cite{singh2021full}), including synthetic text images. It is interesting to highlight that our proposed lite transformer model achieves almost the same performance with a lower quantity of training  data and fewer model parameters. Furthermore, using the same data, our model takes 569 seconds per epoch compared to 940 and  862 seconds per epoch when trained with  \cite{singh2021full,coquenet2022dan} respectively.
}


 Next, we present our results on the RIMES dataset in comparison to the state-of-the-art approaches in Table \ref{rimes}. It can be seen that models from \cite{louradour2014curriculum,coquenet2022end,puigcerver2017multidimensional,coquenet2022dan} require the segmentation bounding boxes annotation to be trained, which is cost-effective. While the models from \cite{coquenet2021span,coquenet2022dan} and the lite transformer are trained on page level to reach a competitive performance without the need for line segmentation.

\begin{table}[H]
\centering
\caption{Evaluation of the lite transformer on the test set of RIMES 2011
dataset  .}\label{rimes}
\begin{tabular}{@{}lll@{}}
\toprule
Type                             & Architecture                                                                 & CER  \\ \midrule
\multirow{3}{*}{Page level} & Coquenet et al \cite{coquenet2021span}                                & 4.17 \\
                                 &{{ Bluche et al \cite{bluche2016joint}}  }                  {}  & {2.9 }\\

                                 & Ours                                                         & 2.9 \\  
                                 &  Coquenet et al \cite{coquenet2022dan}                                  & 1.82 \\
 \\                                
\multirow{3}{*}{Line level}    
                                & Coquenet et al \cite{coquenet2022end}                                  & 3.04 \\
                                 
                                 &  Coquenet et al \cite{coquenet2022dan}                                 & 2.63 \\
                                 & Puigcerver et al \cite{puigcerver2017multidimensional}             & 2.3  \\
                      
                       \bottomrule 
\end{tabular}
\end{table}

Finally, we present in Table \ref{khatt} the results on the KHATT dataset. Compared to models that require segmentation at line level \cite{ahmad2017khatt,ahmad2020deep,noubigh2019contribution}, our model achieves the best performance with a CER of $13.48\%$. This result confirms the ability of our model to recognize other handwritten scripts  by applying a simple transfer-learning process using only page-level labeled images.

\begin{table}[H]
\centering

\caption{Evaluation of the lite transformer on the test set of KHATT dataset.}\label{khatt}
\begin{tabular}{@{}lll@{}}
\toprule
Type                         & Architecture                                 & CER   \\ \midrule
\multirow{3}{*}{Line level} 
                             & DeepKHATT \cite{ahmad2017khatt} & 24.2 \\
                             & DeepLearnig based Arabic script \cite{ahmad2020deep}        & 19.98 \\
                             & CNN+BLSTM \cite{noubigh2019contribution} & 15.8 \\
                             &                                              &       \\
Page level                    & Ours                                         &  13.48  \\ \bottomrule
\end{tabular}
\end{table}

\section{Conclusion}\label{sec7}
In this paper, we proposed MSdocTr-Lite, a lite transformer model for  full-page multi-script handwriting recognition. To solve the common
problem of data scarcity,  MSdocTr-Lite can be trained
on a reasonable amount of data thanks to its light architecture. Contrary to the existing segmentation-based approaches, the
model can be easily fine-tuned on different scripts without using any segmentation label thanks to a simple transfer learning process.

Interesting results have been obtained by MSdocTr-Lite   on English, French, Arabic, and Spanish handwritten scripts. For future research, we intend to integrate a self-supervised training stage, so that, the model can benefit from unlabeled page-level handwriting documents to reduce the annotation cost. We will also explore the capacity of our model to recognize {mixed-script} handwritten documents where many scripts co-exist in the same line/document.  





\end{document}